\documentclass[letterpaper, 10 pt, conference]{ieeeconf} 

\IEEEoverridecommandlockouts

\usepackage{graphics} 
\usepackage{epsfig} 
\usepackage{amssymb}  
\usepackage{subfigure}
\usepackage{cite} 
\usepackage{amsmath}

\graphicspath{ {./img/} }

\newcommand{\Rho}{\mathrm{P}}

\title{\LARGE \bf A Multi-modal Approach to Continuous Material Identification through Tactile Sensing}

\author{A. G\'omez Egu\'{\i}luz$^{1}$ and I. Ra\~n\'o$^{1}$ and %
  S.A. Coleman$^{1}$ and T.M. McGinnity$^{2}$%
  \thanks{$^{1}$A. G\'omez Egu\'{\i}luz, I. Ra\~n\'o, and  S.A. Coleman %
    are with the Intelligent Systems Research Centre, Ulster University, %
    Northern Ireland, UK %
    {\tt\small Gomez\_Eguiluz-A@email.ulster.ac.uk %
      \{i.rano, sa.coleman\}@ulster.ac.uk}}%
  \thanks{$^{2}$ T.M. Mcginnity is with the School of Science \& Technology, %
    Nottingham Trent University, UK %
    {\tt\small martin.mcginnity@ntu.ac.uk}}%
}

\begin{document}

\maketitle
\thispagestyle{empty}
\pagestyle{empty}
\begin{abstract}
Tactile sensing has recently been used in robotics for object
identification, grasping, and material recognition.  Most material
recognition approaches use vibration information from a tactile 
exploration, typically above one second long, to identify the
material. This work proposes a tactile multi-modal (vibration and
thermal) material identification approach based on recursive Bayesian
estimation. Through the frequency response of the vibration induced by
the material and thermal features, like an estimate of the thermal
power loss of the finger, we show that it is possible to identify
materials in less than half a second. Moreover, a comparison between
the use of vibration only and multi-modal identification shows that both
recognition time and classification errors are reduced by adding
thermal information.
\end{abstract}

\section{Introduction}

Tactile sensing has recently attracted significant research interest in
robotics, as a powerful way of enhancing manipulation and
identification. Touch lies at the core of many human skills like
grasping, material identification, and temperature detection, among
others. Similarly, efforts to endow robots with tactile sensing have
led to successful applications of grasping \cite{Gorges.2010} and
palpation \cite{Scheneider.2009} based object classification, object
state identification \cite{Chitta.2011}, grasping improvement and
adaptation \cite{Romano.2011} \cite{hang14hierarchical}, and material
identification
\cite{Edwards.2008,Sinapov.2011,Jamali.2011,Decherchi.2011,Oddo.2011,Dallaire.2013,Chathuranga.2013,Chathuranga.2015,Fishel.2012,Xu.2013,Su.2012,Kerr.2014,Bhattacharjee.2015}.
This work specifically focuses on sequential material identification
using multi-modal -- vibration and heat -- sensing. While most of the
state of the art approaches to material identification through tactile
sensing use only vibration signals, induced when the finger slides
over the material, our experimental results show that the
recognition accuracy can be enhanced using thermal
information. Moreover, because our approach relies on recursive
Bayesian estimation using short windows of readings, a faster material
identification -- compared to other methods -- is achieved.  Thermal
information also contributes to a faster material identification.

Recent research has tackled the material identification problem using
vibration information obtained from different types of tactile
sensors.  Pioneering work in tactile surface recognition
\cite{Edwards.2008} used a custom built sensor with a microphone to detect
vibration signals induced by textured surfaces. The authors defined a
set of features to characterise the vibration signal (modal frequency
and power, and average vibration amplitude). Then, they compared the
classification performance using the \textit{k}-Nearest Neighbour(\textit{k}NN) and the
K-means algorithms with the features sets and Fast
Fourier Transform (FFT) coefficients, projected through Principal
Component Analysis (PCA), as inputs. Their work established a
methodology to classify textures, but it was not evaluated using real
materials.  Another material classification approach used an accelerometer based vibrotactile
sensor \cite{Sinapov.2011}. Change in the acceleration vector (jerk) was used to create
spectrotemporal histograms as features for classification.  Support
Vector Machine (SVM) and \textit{k}NN were used to classify 20 materials with the
features generated from several vibration readings for each
material. The work in \cite{Jamali.2011} presents a tactile sensor
capable of measuring the strain applied on the finger surface. After
preprocessing the input signal, using segmentation, average
removal and band-pass filtering, the authors extract a set of
features consisting of five peaks on the smoothed FFT. Combining the
peak locations with the average strain readings, the authors compare
different machine learning approaches that can successfully differentiate
between nine materials.

In another comparison of machine learning algorithms for vibration
based material identification \cite{Decherchi.2011}, two kernel
methods (SVM and Regularized Least Square), and one neural network
were used to classify pairs of materials based on the raw strain
measurements of the sensor. Although the authors concluded that the SVM
showed the best trade-off between classification accuracy and
computational complexity, the raw sensor signals provided poor
discrimination performance compared with other works. In
\cite{Oddo.2011} a tactile micro-sensor, which can differentiate
surfaces with spatial periods within a $40$~$\mu m$ difference, is
presented and used to classify textiles. A robot finger slid across
the materials for two seconds, and features obtained using wavelet
transforms were used in a \textit{k}NN classifier. Through a set of temporal
domain features computed from a one second signal, the work in
\cite{Dallaire.2013} presents a SVM based classification of material
texture. The sensor consists of a 3D accelerometer, and the feature
vector components are whitened individually before feeding the
classifier.  Another accelerometer based texture recognition system is
presented in \cite{Chathuranga.2013}, where the authors classify seven
different fabrics based on a mixture of temporal (acceleration
variance) and frequency (power spectra) features.  As the
textures cannot be distinguished using these features, the authors
train a neural network using the FFT coefficients over a given frequency
range.  Recently, \cite{Chathuranga.2015} explored real time
classification of eight materials using a soft three axis tactile
sensor. The vibration mean value and the Frobenius norm of its
covariance matrix were fed into a cascade of binary SVMs, achieving
$89\%$ identification accuracy.  

All these pioneering works are based on custom made sensors to detect
the vibration induced by a surface texture. The development of the
SynTouch BioTAC fingertip which provides multiple types of tactile
information opened a window of opportunity to investigate multi-modal
surface recognition.  A series of works \cite{Fishel.2012}
\cite{Xu.2013} \cite{Su.2012}, implementing human based Bayesian
exploratory movements, classify materials using the BioTAC
vibration, impedance, and temperature sensors. The authors identify
surfaces using a combination of features measuring vibration and the
friction force between the materials and the finger. If the
identification certainty was not high enough the authors introduced
active sensing using Bayesian inference to perform a different
exploratory movement. Another multi-modal approach to material
identification is presented in \cite{Kerr.2014} based on readings of
temperature and vibration projected through PCA. The authors show that
an artificial neural network outperforms humans in similar
experimental conditions. The heat transfer from a custom tactile
sensor to an object is used for material identification in
\cite{Bhattacharjee.2015}. The authors showed that, although changes
in the experiment set-up (i.e. initial conditions, ambient temperature
and contact duration) have an impact on the performance, a multi-class
SVM classifies 11 materials with $98\%$ accuracy in 1.5 seconds of
contact.

All these works achieve good material identification performance, however most of them do not exploit multi-modal tactile information. Moreover, they require long sequential readings, typically one second.  The contribution of this work is two-fold.
First, we use a recursive estimation approach with short tactile
readings, which allows fast, under $0.5$ secs on average, and
very accurate material identification using vibration signals.
Second, we show that including temperature information significantly
reduces the time needed to identify the correct material and the
number of classification errors.  Our approach generates sequences of
estimates of the posterior material probabilities instead of a single
decision, and the material with the highest probability is considered
the right one.  The rest of the paper is organised as follows.
Section~\ref{sect:RBE} presents the methodology used for the proposed
online material identification approach, including the feature
generation from the raw data, probabilistic modelling, and the
recursive classification technique. Experimental results comparing
vibration and multi-modal material identification approaches are
presented in Section~\ref{sect:Experiments}.
Section~\ref{sect:conclusion} concludes the paper and highlights
future research steps.

\section{Multi-modal Recursive Material Identification}
\label{sect:RBE}

We use the SynTouch BioTAC \cite{BiotacManual} finger tip as our
experimental platform. It provides a variety of sensing modalities,
pressure, vibration, temperature, heat flux and finger skin
deformation, with different frequencies. The finger consists of a
rigid core where the sensors are located, covered by a wrinkled
rubber skin. The core has a heating system and a thermistor to read
the temperature of the finger.  A gel separates the core from the
skin. An array of impedance sensors at the core measure the
deformation of the skin in contact with the materials. The rest of
this section will present datails on how the vibration and thermal
information is processed in our recursive Bayesian estimation framework.  In a
nutshell, our approach processes the vibration and thermal flux
(together with the impedance) signals to obtain sets of features
modelled as mixtures of Gaussians for each material.  The recursive
Bayesian estimation algorithm uses these models for continuous
identification of the probability of each material.

\subsection{Vibration Signal Processing}
\label{sec:FFT}

When the BioTAC slides over a surface, the interaction with the rubber
wrinkles produces vibration in the skin which is transferred to the
fluid, and measured by the pressure sensor. This signal is low pass
filtered to generate a pressure measurement, and band pass filtered to
generate a vibration signal.  The vibration induced by the interaction
of the wrinkled rubber skin and the texture of the materials can be
seen as a combination of oscillatory signals with the frequency
spectrum dependent on the material. Instead of defining features in
the frequency or temporal domains, we directly use the
Fourier Transform (FT) of the vibration signal to characterise the
vibration response induced by the material texture. Specifically, the
Fast Fourier Transform (FFT) algorithm was used to convert the
vibration signal, $\rho(t)$ into the frequency domain
$\rho(\omega)$. According to the existing literature
\cite{Edwards.2008} good discrimination results are obtained by
restricting the FFT to a range of frequencies between $2$~$Hz$ and
$500$~$Hz$. It is worth stating that we compute the FFT for small ($0.25$ secs)
non-overlapping windows of the signal of duration $\Delta t$ , in order
to perform online identification.

The vibration FFT $\rho(w)$ is a high dimensional vector (dimension
$d_{\Delta t}$) of complex numbers where $d_{\Delta t}$ depends on the
sampling period and the selected time interval $\Delta t$.  To obtain
a feature vector $\bar{\rho}\in\Re^d$ with a lower dimension,
Principal Component Analysis (PCA) was applied to $\rho(\omega)$,
where $\bar{\rho}$ denotes the projection of $\rho(\omega)$ and
$d\ll d_{\Delta t}$.  Because $\rho(\omega)$ is complex, yet the
relevant information to classify the materials is in its modulus, the
principal components were obtained from the modulus of the FFT. We
observed, however, that better discrimination results were obtained if
the centering process was performed in the complex FFT space, i.e. we
applied PCA to the modulus upon centering with the complex mean.

\subsection{Thermal features}
\label{sec:Temperature}

The BioTAC sensor has a heating device and measures the temperature at
the core and the heat flux leaving the finger. The thermal energy lost
depends on the temperature difference between the finger and the
external material, the contact area, and the thermal conductivity of
the material. In fact, the thermal energy lost per unit of time
(thermal power $\frac{\partial E}{\partial T}$) is the integral of the
heat flux over the contact surface:

\begin{eqnarray}
\frac{\partial E}{\partial T} = \oint_S{\vec{\phi}\cdot\vec{dS}}
\end{eqnarray}

where $\vec{\phi}$ is the heat flux, and the integral is computed on
the contact area $S$. As in our experiments the BioTAC is warmer than the material,
the flux $\vec{\phi}$ always leaves the finger, and its modulus,
measured by the finger, increases with the temperature difference
and the thermal conductivity of the material. Assuming that most of
the energy is lost because of the temperature difference with the
material (friction forces are too weak to generate enough thermal
energy and the energy lost through the air is small), the thermal flux
will be directed towards the surface normal. Moreover, the contact
area is typically small, so we can approximate the power loss as the
product of the average flux modulus $\bar{\phi}$ by the contact
area. If we assume all objects to be identified are at the same (room)
temperature, the temperature difference can be measured as the
temperature at the BioTAC core, and therefore $\frac{\bar{\phi}A}{T}$
is a measure of the thermal conductivity of the material, where $A$ is
the contact area, and $T$ is the finger core temperature.

To compute the contact area we use the 19 electrodes placed in the
core under the BioTAC's skin. The electrodes measure impedance which
is related to the distance between the core and the rubber skin at
their corresponding locations. Upon contact, the skin deformation makes
the readings in the closest electrodes decrease, and, therefore,
electrodes with a negative value w.r.t. their resting level indicate
contact. We approximate the contact area of each electrode $i$ as a
circle of radius $r_i$ equal to half the distance between the
electrode and its nearest neighbour. Hence, we compute the contact
area of the fingertip as a weighted average of these areas:

\begin{eqnarray}
 A=\sum_{i}^{}\lambda_i\pi r^2_i 
\end{eqnarray}

where $\lambda_i\in[0,1]$ is an scale factor that depends on the value
measured by each electrode. The scaling factor $\lambda_i$ is a
piece-wise linear function of the average impedance value $\bar{e}_i$
of each electrode during the time interval $\Delta t$, such that at
the resting level (or above) $\lambda_i$ is zero, and it increases to
$1$ for decreasing impedances up to a fixed minimum threshold $e_m$
(in our case $e_m=-400$), and is $1$ for values below that threshold.
Besides using the power loss per unit of temperature, just described,
we defined two other features based on heat flux. We perform a
linear regression of the thermal flux as a function of time in the
$\Delta t$ interval, and use the slope and the regression error as
additional features, as we experimentally found they help for material
identification. Therefore, we obtained a three dimensional heat based
feature vector $\theta$ for each $\Delta t$ time window.

\subsection{Material Recursive Bayesian Estimation}

We will denote by $M$ the discrete random variable encoding the $N$
materials to be identified, i.e. $\{m^1,m^2,\cdots,m^N\}$,
$\Rho\in\Re^d$ is the random vector of vibration features (FFT
projected through PCA) and $\Theta\in\Re^3$ the vector thermal
features defined in section~\ref{sec:Temperature}.  We estimate the
respective likelihood functions of the feature vectors for each
material, $p(\Rho=\bar{\rho}|M=m^j)$ and $p(\Theta=\theta|M=m^j)$,
using mixtures of Gaussian distributions, i.e.  Gaussian Mixture
Models (GMM). The parameters of the GMMs are obtained using the
Expectation-Maximisation (EM) algorithm, and the number of Gaussians
was chosen for each material and each sensor modality based on the
decrease of the likelihood of the training data, i.e. increasing the
number of components to find the point at which the growth in the
likelihood started to slow down.

At this stage we have models of the likelihood functions
$p(\bar{\rho}|m^j)$ and $p(\theta|m^j)$ for the materials, which would
already allow identification of the materials using a maximum likelihood
(ML) classifier. Given a set of prior material probabilities $p(m^j)$,
one could also estimate through the Bayes rule the posterior material
probabilities $p(m^j|\bar{\rho},\theta)$, and classify according to
the maximum \textit{a posteriori} (MAP) probability.  Under the
assumption of uninformative priors $p(m^j)=\frac{1}{N}$ for all $j$,
the classification result of ML and MAP approaches are identical.
However, as already stated, our aim is to identify materials online
without the need to collect a long sequence of readings. Therefore we
will use the feature vectors $\bar{\rho}$ and $\theta$ to iteratively
obtain new probability posteriors for each material. The material
prior $p(m^j_{k+1})$ at time step $k+1$ will simply be the posterior
from the previous iteration $p(m^j_k|\bar{\rho}_{k},\theta_{k})$.

When a material is presented to the robot hand for identification, the
initial prior probabilities are distributed evenly among all
materials.  We obtain from the BioTAC data stream the vibration and
thermal feature vectors, $\bar{\rho}_k$ and $\theta_k$, in windows of
time $\Delta t$, and update the material probabilities using:

\begin{eqnarray}
  p(m^j_{k}|\bar{\rho}_{k},\theta_k) &=& \frac{p(\bar{\rho}_k,\theta_k|m^j_k)
    p(m^j_{k-1}|\bar{\rho}_{k-1},\theta_{k-1})} {p(\bar{\rho}_k,\theta_k)}
       \label{eq:RBE2}
\end{eqnarray}

where the normalisation constant $p(\bar{\rho}_k,\theta_k)$ can be
obtained as:

\begin{eqnarray}
  p(\bar{\rho}_k,\theta_k)&=&
  \sum_i^Np(\bar{\rho}_k,\theta_k|m^i_k)p(m^i_{k-1}|\bar{\rho}_{k-1},\theta_{k-1})
  \label{eq:RBE3}
\end{eqnarray}

and we assume the vibration and thermal features are conditionally
independent
$p(\bar{\rho}_k,\theta_k|m^j_k)=p(\bar{\rho}_k|m^j_k)p(\theta_k|m^j_k)$,
with each individual likelihood function given by the corresponding
GMM model for material $m^j$.

In each iteration the algorithm updates the posterior probability
$p(m^{j}_{k}|\bar{\rho}_{k})$ for all materials, $j=1,\cdots, N$, and
the material with the highest posterior can be considered to be the
one presented to the robot. Alternatively, a confidence level could be
defined to decide for a material only if the posterior probability is
above some threshold.  Instead of predicting the perceived texture in
one episode, the recursive Bayesian estimation algorithm incrementally
updates the probability estimate of the material.  As a baseline to
compare the recognition accuracy of our multi-modal approach, we first
apply the recursive Bayesian estimation procedure to the vibration
information (see section~\ref{sect:Experiments}) using expressions
(\ref{eq:RBE2}) and (\ref{eq:RBE3}), without the thermal features
$\theta$.

\section{Experimental Results}
\label{sect:Experiments}

Our experimental setup consists of a turntable moved by a step motor
through a set of reduction gears. The motor is controlled by an
Arduino board running code to set the turning speed. The fingertip is
attached to a worm drive bar moved up and down by a second motor also
controlled by the Arduino board. If the bar is driven down, the BioTAC
touches the material on the turn table. To gather the training and
testing data-sets, we first set the speed of the turntable and moved
the fingertip down until it touched the surface. After a few seconds
we collect readings, continuously storing all the information provided
by the BioTAC sensor running at a sampling rate of $4.4$~$KHz$. This
is the bandwidth at which the sensor provides individual readings, yet
readings for pressure, temperature and impedance, have lower
frequencies. For instance the vibration signal has a sampling
frequency of $2.2$~$KHz$, while pressure, absolute temperature, and
thermal flux have frequencies of just $100$~$Hz$.

\begin{table}[]
  \centering
  \begin{tabular}{|r|l|l|r|l|}			
    \cline{1-2} \cline{4-5}
    \textbf{Idx} & \textbf{Material} &  & 
    \textbf{Idx} & \textbf{Material} \\ 
    \cline{1-2} \cline{4-5} 
    1  & Synthetic Green fabric &  & 18 & Genuine leather  \\ 
    \cline{1-2} \cline{4-5} 
    2  & Synthetic Pink fabric 1 &  & 19 & Linen  \\ 
    \cline{1-2} \cline{4-5} 
    3  & Synthetic Pink fabric 2 &  & 20 & Mirror  \\ 
    \cline{1-2} \cline{4-5} 
    4  & Cardboard box &  & 21 & Normal paper \\ 
    \cline{1-2} \cline{4-5} 
    5  & Cardboard disk &  & 22 & Ping pong paddle 1 \\ 
    \cline{1-2} \cline{4-5} 
    6  & Carpet &  & 23 & Ping pong paddle 2 \\ 
    \cline{1-2} \cline{4-5} 
    7  & Rubber &  & 24 & Plastic \\ 
    \cline{1-2} \cline{4-5} 
    8  & Baize &  & 25 & Plastic dish \\ 
    \cline{1-2} \cline{4-5} 
    9  & Can of drink &  & 26 & Rough fabric \\ 
    \cline{1-2} \cline{4-5} 
    10 & Copper &  & 27& Slate stone \\ 
    \cline{1-2} \cline{4-5}
    11 & Cork &  & 28 & Sponge 1 \\ 
    \cline{1-2} \cline{4-5} 
    12 & 100\% Cotton &  & 29 & Sponge 2 \\ 
    \cline{1-2} \cline{4-5} 
    13 & Padded envelope 1&  & 30 & Leopard fabric 1 \\ 
    \cline{1-2} \cline{4-5} 
    14 & Padded envelope 2 &  & 31 & Leopard fabric 2 \\ 
    \cline{1-2} \cline{4-5}
    15 & Aluminium &  & 32  & Watercolour paper \\ 
    \cline{1-2} \cline{4-5} 
    16 & Synthetic leather &  & 33 & Wood \\ 
    \cline{1-2} \cline{4-5} 
    17 & Floor tiles &  & 34 & Peach skin fabric \\ 
    \cline{1-2} \cline{4-5}
  \end{tabular}
  \caption{List of materials used in the classification experiments}
  \label{table:Materials}
\end{table}

\begin{figure}[]
  \centering
  \includegraphics[width=0.8\columnwidth]{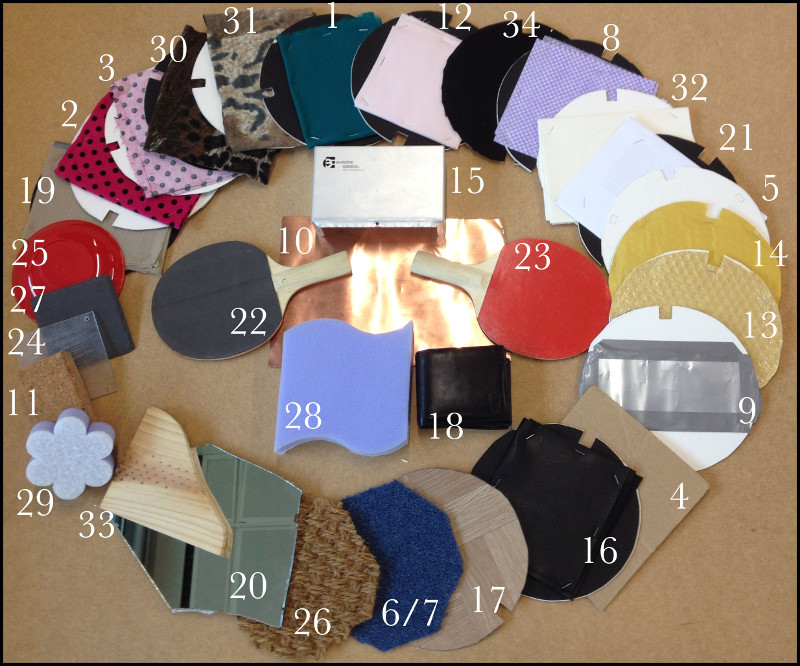}
  \caption{Materials used in the experiments}
  \label{fig:allMaterials}
\end{figure}

The materials to be identified were purposely selected to include
different groups (metals, plastic, textiles\ldots) and to contain
similar textures.  Table~\ref{table:Materials} presents a list of the
materials used in this work, also shown in
Figure~\ref{fig:allMaterials}.  It can be seen that the data set
contains materials with very different textures such as cork and a
glass mirror, but also some fabrics and materials which are alike. To
increase the number of materials, some of the fabrics were used from
both sides of the cloth (materials $2$ and $3$, and materials $30$ and
$31$). We also aimed to differentiate between genuine and synthetic
leather represented by materials $16$ and $18$, while materials $13$
and $14$ were obtained using both sides of a padded envelope, $13$
being the side of the bubbles, i.e. plastic, and $14$ the paper side.
Other pairs correspond to both sides of the same object such as
materials $6$-$7$, $22$-$23$ and $28$-$29$, although their surfaces
were clearly different.

\subsection{Vibration based material identification}
\label{subsect:single-material}

In our first experiment we classified the 34 different textures listed
in Table~\ref{table:Materials} using only vibration information and
the approach presented in Section~\ref{sect:RBE}.  The time domain
vibration signal was split into non-overlapping windows of $\Delta
t=0.25$ seconds, which is therefore the minimum recognition time, and
sets the frame rate for the continuous material classification to
$4$~$Hz$.  It is worth noting that, given the sampling rate and the
short time interval selected ($0.25$~$sec$), the frequency interval
between spectral components is approximately $4$~$Hz$, that is our real
minimum frequency in the band pass filter.  This window corresponds to
a sequence of $550$ pressure samples, and upon computing the FFT and
keeping the selected range of frequencies ($4-500$~$Hz$) the resulting
number of spectral components was $124$. Increasing $\Delta t$ would
provide a finer FFT approximation at the cost of reducing the
classification frequency. As stated in Section~\ref{sec:FFT} we
reduced the $124$ spectral components to a $16$ dimensional space,
computing the complex average of the training data-set for all 34
materials, and projecting their amplitude vectors through PCA,
therefore $\bar{\rho}\in\Re^{16}$. The dimension of the reduced
features was chosen to keep $97\%$ of the total variance of the
training data-set.

To evaluate our proposed approach we performed 10-fold cross validation on 5
minutes of data sequences recorded for each material. We split the
data set into 10 groups of 30 seconds of sequential readings using
alternatively 9 of the groups for training and one for
testing. As our approach classifies the materials sequentially we
used two performance measurements in the test data sequences,
namely recognition errors and recognition time. For each one of the
10-fold cross-validation procedures, once the projection matrix and
GMMs were obtained, the evaluation was conducted with the remaining $30$
seconds time series for each material.  Assuming uninformative priors
and starting from the first sample the full sequence was processed,
and the number of misclassifications and iterations required to
successfully identify the material was stored.  The first $\Delta t$
readings at the beginning of the sequence were then discarded
(creating a shorter test sequence) and the process was repeated until
the sequence was only of length $\Delta t$ (i.e. $0.25$ seconds).
This process was repeated for all the materials in the 10 testing
sequences, and the average number of iterations required to identify each
material was computed. It was considered that a material was
successfully identified when the maximum posterior probability
$p(m^{j}_k|\bar{\rho}_k)$ for that material was the highest.

As for the missclasification rate, we found that the system always
properly identifies the materials when enough evidence was collected,
resulting in a perfect recognition rate. However, this process can
take several iterations, and, at the end of the testing sequences, the
algorithm was sometimes unable to accumulate enough evidence to
properly classify the textures. This leads to an average error rate
of $1.21\%$ with the errors mainly occurring in the first iteration
(i.e. maximum likelihood classification).

For our second performance measurement, the recognition time, the
number of iterations needed to successfully identify the material was
translated into seconds multiplying by $0.25$ ($\Delta t$).
Table~\ref{table:10FoldCV_time} shows the average recognition times
and standard deviations for each material over the 10-fold
cross-validation process.  As it can be seen that the vibration based
recursive Bayesian estimation identifies materials within less than
$0.5$ seconds, and the total average time across materials (last
column of the table) is approximately $1.5$ iterations.

\begin{table}[h]
	\centering
	\begin{tabular}{llllllll}
		\hline
		\multicolumn{1}{|l|}{\textbf{Material}} & \multicolumn{1}{c|}{1}    & \multicolumn{1}{c|}{2}    & \multicolumn{1}{c|}{3}    & \multicolumn{1}{c|}{4}    & \multicolumn{1}{c|}{5}    & \multicolumn{1}{c|}{6}    & \multicolumn{1}{c|}{7}             \\ \hline
		\multicolumn{1}{|l|}{\textbf{mean (secs)}}  & \multicolumn{1}{l|}{0.39}  & \multicolumn{1}{l|}{0.35} & \multicolumn{1}{l|}{0.35} & \multicolumn{1}{l|}{0.43} & \multicolumn{1}{l|}{0.39}  & \multicolumn{1}{l|}{0.3} & \multicolumn{1}{l|}{0.33}          \\ \hline
		\multicolumn{1}{|l|}{\textbf{std.dev. (secs)}}  & \multicolumn{1}{l|}{0.21}  & \multicolumn{1}{l|}{0.19} & \multicolumn{1}{l|}{0.19} & \multicolumn{1}{l|}{0.28} & \multicolumn{1}{l|}{0.25}  & \multicolumn{1}{l|}{0.12} & \multicolumn{1}{l|}{0.14}          \\ \hline
		\hline
		\multicolumn{1}{|l|}{\textbf{Material}} & \multicolumn{1}{c|}{8}    & \multicolumn{1}{c|}{9}    & \multicolumn{1}{c|}{10}   & \multicolumn{1}{c|}{11}   & \multicolumn{1}{c|}{12}   & \multicolumn{1}{c|}{13}   & \multicolumn{1}{c|}{14}            \\ \hline
		\multicolumn{1}{|l|}{\textbf{mean (secs)}} & \multicolumn{1}{l|}{0.44} & \multicolumn{1}{l|}{0.34} & \multicolumn{1}{l|}{0.3} & \multicolumn{1}{l|}{0.3} & \multicolumn{1}{l|}{0.35} & \multicolumn{1}{l|}{0.33} & \multicolumn{1}{l|}{0.34}          \\ \hline
		\multicolumn{1}{|l|}{\textbf{std.dev. (secs)}}  & \multicolumn{1}{l|}{0.26}  & \multicolumn{1}{l|}{0.16} & \multicolumn{1}{l|}{0.2} & \multicolumn{1}{l|}{0.11} & \multicolumn{1}{l|}{0.18}  & \multicolumn{1}{l|}{0.16} & \multicolumn{1}{l|}{0.17}          \\ \hline
		\hline
		\multicolumn{1}{|l|}{\textbf{Material}} & \multicolumn{1}{c|}{15}   & \multicolumn{1}{c|}{16}   & \multicolumn{1}{c|}{17}   & \multicolumn{1}{c|}{18}   & \multicolumn{1}{c|}{19}   & \multicolumn{1}{c|}{20}   & \multicolumn{1}{c|}{21}            \\ \hline
		\multicolumn{1}{|l|}{\textbf{mean (secs)}}  & \multicolumn{1}{l|}{0.3} & \multicolumn{1}{l|}{0.44} & \multicolumn{1}{l|}{0.46} & \multicolumn{1}{l|}{0.32} & \multicolumn{1}{l|}{0.49} & \multicolumn{1}{l|}{0.29}  & \multicolumn{1}{l|}{0.39}           \\ \hline
		\multicolumn{1}{|l|}{\textbf{std.dev. (secs)}}  & \multicolumn{1}{l|}{0.16}  & \multicolumn{1}{l|}{0.25} & \multicolumn{1}{l|}{0.34} & \multicolumn{1}{l|}{0.14} & \multicolumn{1}{l|}{0.35}  & \multicolumn{1}{l|}{0.11} & \multicolumn{1}{l|}{0.25}          \\ \hline
		\hline
		\multicolumn{1}{|l|}{\textbf{Material}} & \multicolumn{1}{c|}{22}   & \multicolumn{1}{c|}{23}   & \multicolumn{1}{c|}{24}   & \multicolumn{1}{c|}{25}   & \multicolumn{1}{c|}{26}   & \multicolumn{1}{c|}{27}   & \multicolumn{1}{c|}{28}            \\ \hline
		\multicolumn{1}{|l|}{\textbf{mean (secs)}}  & \multicolumn{1}{l|}{0.38} & \multicolumn{1}{l|}{0.36} & \multicolumn{1}{l|}{0.38} & \multicolumn{1}{l|}{0.38} & \multicolumn{1}{l|}{0.27} & \multicolumn{1}{l|}{0.42} & \multicolumn{1}{l|}{0.29}          \\ \hline
		\multicolumn{1}{|l|}{\textbf{std.dev (secs)}}  & \multicolumn{1}{l|}{0.22}  & \multicolumn{1}{l|}{0.19} & \multicolumn{1}{l|}{0.22} & \multicolumn{1}{l|}{0.21} & \multicolumn{1}{l|}{0.07}  & \multicolumn{1}{l|}{0.25} & \multicolumn{1}{l|}{0.13}          \\ \hline
		\hline
		\multicolumn{1}{|l|}{\textbf{Material}} & \multicolumn{1}{c|}{29}   & \multicolumn{1}{c|}{30}   & \multicolumn{1}{c|}{31}   & \multicolumn{1}{c|}{32}   & \multicolumn{1}{c|}{33}   & \multicolumn{1}{c|}{34}   & \multicolumn{1}{c|}{{\bf Avg.}} \\ \hline
		\multicolumn{1}{|l|}{\textbf{mean (secs)}} & \multicolumn{1}{l|}{0.32} & \multicolumn{1}{l|}{0.27} & \multicolumn{1}{l|}{0.27} & \multicolumn{1}{l|}{0.46} & \multicolumn{1}{l|}{0.52} & \multicolumn{1}{l|}{0.4} & \multicolumn{1}{l|}{{\bf 0.36}}   \\ \hline
		\multicolumn{1}{|l|}{\textbf{std.dev (secs)}}  & \multicolumn{1}{l|}{0.14}  & \multicolumn{1}{l|}{0.08} & \multicolumn{1}{l|}{0.08} & \multicolumn{1}{l|}{0.3} & \multicolumn{1}{l|}{0.34}  & \multicolumn{1}{l|}{0.26} & \multicolumn{1}{l|}{{\bf 0.2}}         \\ \hline
	\end{tabular}
	\caption{Average time needed for material classification using 10-fold cross validation}
	\label{table:10FoldCV_time}
\end{table}

\begin{figure}[]
  \begin{center}
    \includegraphics[width=0.9\columnwidth]{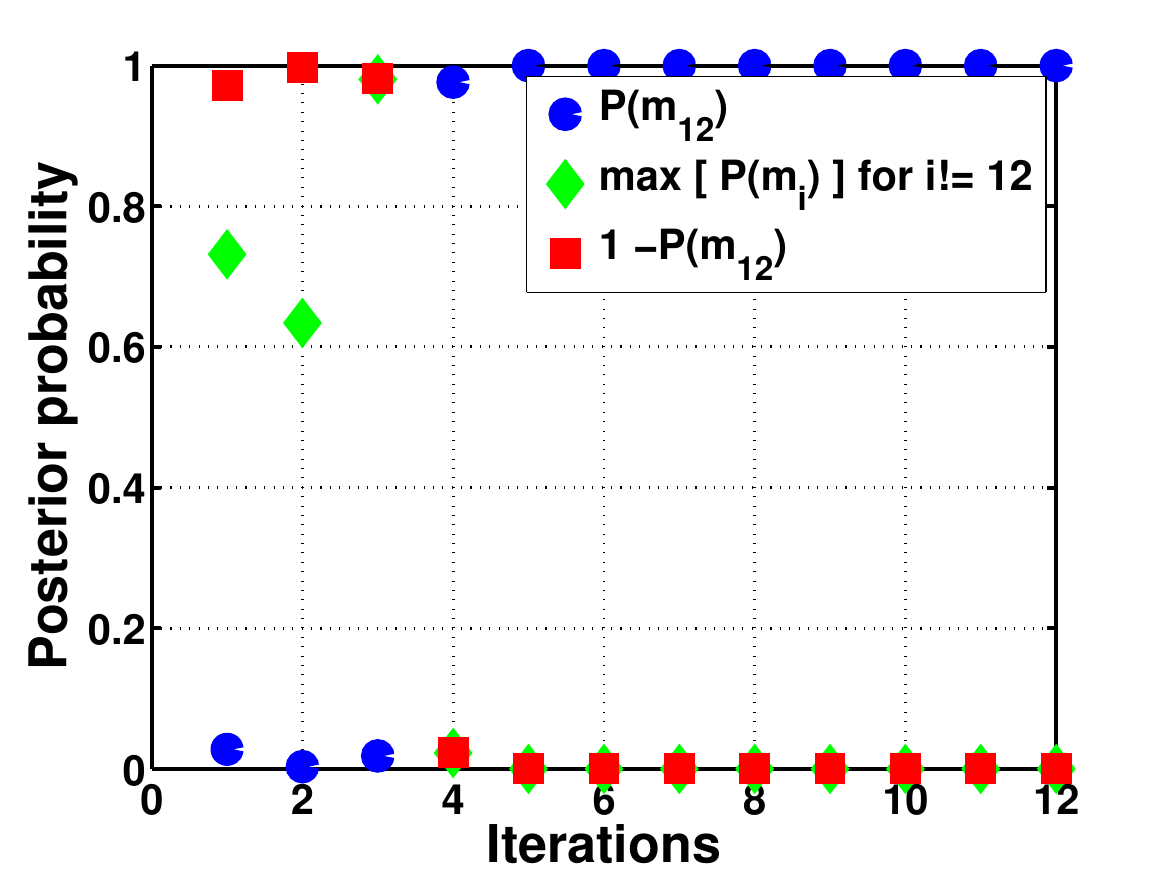}
  \caption{Iterative estimation of material 12 using vibration}
  \label{figure:RBE}
  \end{center}
\end{figure}

Figure~\ref{figure:RBE} shows an example of the vibration based
recursive estimation during 12 iterations (3 seconds) of test data for
material 12 (cotton). The circle corresponds to the posterior
probability of the correct material, the square corresponds to the combined
probability of all other materials (i.e. excluding cotton), and the
diamond corresponds to the second most likely material. The figure shows that
correct identification occurs after four iterations ($1$
second). During the first two iterations there is not clear certainty
on the material, in the third the posterior of material 12 is rather
low. However, after the fourth iteration the classification is
correct.

\subsection{Improving recognition through thermal features} 
\label{subsect:temperature}

The results presented in Section~\ref{subsect:single-material} are our
baseline to compare the multi-modal tactile approach to material
identification. Therefore, in this section, we perform an identical
evaluation procedure including the vibration and thermal features
($\theta_k$) described in section~\ref{sec:Temperature}. As already
stated, the sampling frequency of the thermal flux, temperature, and
impedance sensors, is lower than the vibration sampling, and our
$\Delta t$ window corresponds to $25$ samples of each of these
signals. To estimate the thermal power loss we averaged the heat flux
and impedance readings during the measuring interval, while the heat
flux slope and error are computed with all the samples.  The number of
Gaussians in the mixture model for the thermal features was typically
$2$, except for materials 4, 8, 12 and 23 which we estimated as $3$
using the changes in the training data likelihood.

\begin{table}[]
	\centering
	\begin{tabular}{llllllll}
		\hline
		\multicolumn{1}{|l|}{\textbf{Material}} & \multicolumn{1}{c|}{1}    & \multicolumn{1}{c|}{2}    & \multicolumn{1}{c|}{3}    & \multicolumn{1}{c|}{4}    & \multicolumn{1}{c|}{5}    & \multicolumn{1}{c|}{6}    & \multicolumn{1}{c|}{7}             \\ \hline
		\multicolumn{1}{|l|}{\textbf{mean (secs)}}  & \multicolumn{1}{l|}{0.28}  & \multicolumn{1}{l|}{0.28} & \multicolumn{1}{l|}{0.29} & \multicolumn{1}{l|}{0.28} & \multicolumn{1}{l|}{0.31}  & \multicolumn{1}{l|}{0.26} & \multicolumn{1}{l|}{0.27}          \\ \hline
		\multicolumn{1}{|l|}{\textbf{std.dev. (secs)}}  & \multicolumn{1}{l|}{0.08}  & \multicolumn{1}{l|}{0.1} & \multicolumn{1}{l|}{0.11} & \multicolumn{1}{l|}{0.1} & \multicolumn{1}{l|}{0.14}  & \multicolumn{1}{l|}{0.06} & \multicolumn{1}{l|}{0.08}         \\ \hline
		\hline
		\multicolumn{1}{|l|}{\textbf{Material}} & \multicolumn{1}{c|}{8}    & \multicolumn{1}{c|}{9}    & \multicolumn{1}{c|}{10}   & \multicolumn{1}{c|}{11}   & \multicolumn{1}{c|}{12}   & \multicolumn{1}{c|}{13}   & \multicolumn{1}{c|}{14}            \\ \hline
		\multicolumn{1}{|l|}{\textbf{mean (secs)}} & \multicolumn{1}{l|}{0.33} & \multicolumn{1}{l|}{0.32} & \multicolumn{1}{l|}{0.26} & \multicolumn{1}{l|}{0.25} & \multicolumn{1}{l|}{0.28} & \multicolumn{1}{l|}{0.29} & \multicolumn{1}{l|}{0.26}          \\ \hline
		\multicolumn{1}{|l|}{\textbf{std.dev. (secs)}}  & \multicolumn{1}{l|}{0.18}  & \multicolumn{1}{l|}{0.17} & \multicolumn{1}{l|}{0.03} & \multicolumn{1}{l|}{0.02} & \multicolumn{1}{l|}{0.09}  & \multicolumn{1}{l|}{0.13} & \multicolumn{1}{l|}{0.06}          \\ \hline
		\hline
		\multicolumn{1}{|l|}{\textbf{Material}} & \multicolumn{1}{c|}{15}   & \multicolumn{1}{c|}{16}   & \multicolumn{1}{c|}{17}   & \multicolumn{1}{c|}{18}   & \multicolumn{1}{c|}{19}   & \multicolumn{1}{c|}{20}   & \multicolumn{1}{c|}{21}            \\ \hline
		\multicolumn{1}{|l|}{\textbf{mean (secs)}}  & \multicolumn{1}{l|}{0.25} & \multicolumn{1}{l|}{0.38} & \multicolumn{1}{l|}{0.32} & \multicolumn{1}{l|}{0.30} & \multicolumn{1}{l|}{0.30} & \multicolumn{1}{l|}{0.26}  & \multicolumn{1}{l|}{0.26}           \\ \hline
		\multicolumn{1}{|l|}{\textbf{std.dev. (secs)}}  & \multicolumn{1}{l|}{0.01}  & \multicolumn{1}{l|}{0.25} & \multicolumn{1}{l|}{0.17} & \multicolumn{1}{l|}{0.14} & \multicolumn{1}{l|}{0.14}  & \multicolumn{1}{l|}{0.5} & \multicolumn{1}{l|}{0.04}          \\ \hline
		\hline
		\multicolumn{1}{|l|}{\textbf{Material}} & \multicolumn{1}{c|}{22}   & \multicolumn{1}{c|}{23}   & \multicolumn{1}{c|}{24}   & \multicolumn{1}{c|}{25}   & \multicolumn{1}{c|}{26}   & \multicolumn{1}{c|}{27}   & \multicolumn{1}{c|}{28}            \\ \hline
		\multicolumn{1}{|l|}{\textbf{mean (secs)}}  & \multicolumn{1}{l|}{0.27} & \multicolumn{1}{l|}{0.29} & \multicolumn{1}{l|}{0.27} & \multicolumn{1}{l|}{0.31} & \multicolumn{1}{l|}{0.25} & \multicolumn{1}{l|}{0.32} & \multicolumn{1}{l|}{0.25}          \\ \hline
		\multicolumn{1}{|l|}{\textbf{std.dev (secs)}}  & \multicolumn{1}{l|}{0.07}  & \multicolumn{1}{l|}{0.11} & \multicolumn{1}{l|}{0.07} & \multicolumn{1}{l|}{0.14} & \multicolumn{1}{l|}{0.01}  & \multicolumn{1}{l|}{0.17} & \multicolumn{1}{l|}{0.01}          \\ \hline
		\hline
		\multicolumn{1}{|l|}{\textbf{Material}} & \multicolumn{1}{c|}{29}   & \multicolumn{1}{c|}{30}   & \multicolumn{1}{c|}{31}   & \multicolumn{1}{c|}{32}   & \multicolumn{1}{c|}{33}   & \multicolumn{1}{c|}{34}   & \multicolumn{1}{c|}{{\bf Avg.}} \\ \hline
		\multicolumn{1}{|l|}{\textbf{mean (secs)}} & \multicolumn{1}{l|}{0.27} & \multicolumn{1}{l|}{0.26} & \multicolumn{1}{l|}{0.25} & \multicolumn{1}{l|}{0.29} & \multicolumn{1}{l|}{0.30} & \multicolumn{1}{l|}{0.26} & \multicolumn{1}{l|}{{\bf 0.28}}   \\ \hline
		\multicolumn{1}{|l|}{\textbf{std.dev (secs)}}  & \multicolumn{1}{l|}{0.07}  & \multicolumn{1}{l|}{0.06} & \multicolumn{1}{l|}{0.01} & \multicolumn{1}{l|}{0.13} & \multicolumn{1}{l|}{0.12}  & \multicolumn{1}{l|}{0.06} & \multicolumn{1}{l|}{{\bf 0.09}}         \\ \hline
	\end{tabular}
	\caption{Temperature for recognition improvement. Average time needed using 10-fold cross validation}
	\label{table:10FoldCV_time_temperature}
\end{table}

Our experiments showed that including thermal information reduces the
average errors on the material classification with respect to
vibration only classification. Specifically, the number of
misclassification samples measured as explained in
Section~\ref{subsect:single-material} becomes $1.02\%$, $15.6\%$
reduction in the classification errors, w.r.t the vibration only
classification described in the previous section.
Table~\ref{table:10FoldCV_time_temperature} shows the average and
standard deviation of the classification time using the multi-modal
approach. Both, the overall average time and the standard deviation
are reduced (cf. Table~\ref{table:10FoldCV_time}) which means the
recognition occurs faster ($22\%$ improvement on the average), but
it is also more stable. Since the average recognition time is now
$0.28$ seconds and the recognition period is $0.25$ seconds there is
not much room for improvement, which actually means almost all of the
materials are now identified in one step. While for some materials
including thermal information slightly improves the recognition time
(e.g. materials 30/31), in most of the cases the improvement in time is
above $10\%$, with material 33 (wood) experiencing a time recognition
improvement of over $40\%$.

\begin{figure}[]
  \begin{center}
    \includegraphics[width=0.9\columnwidth]{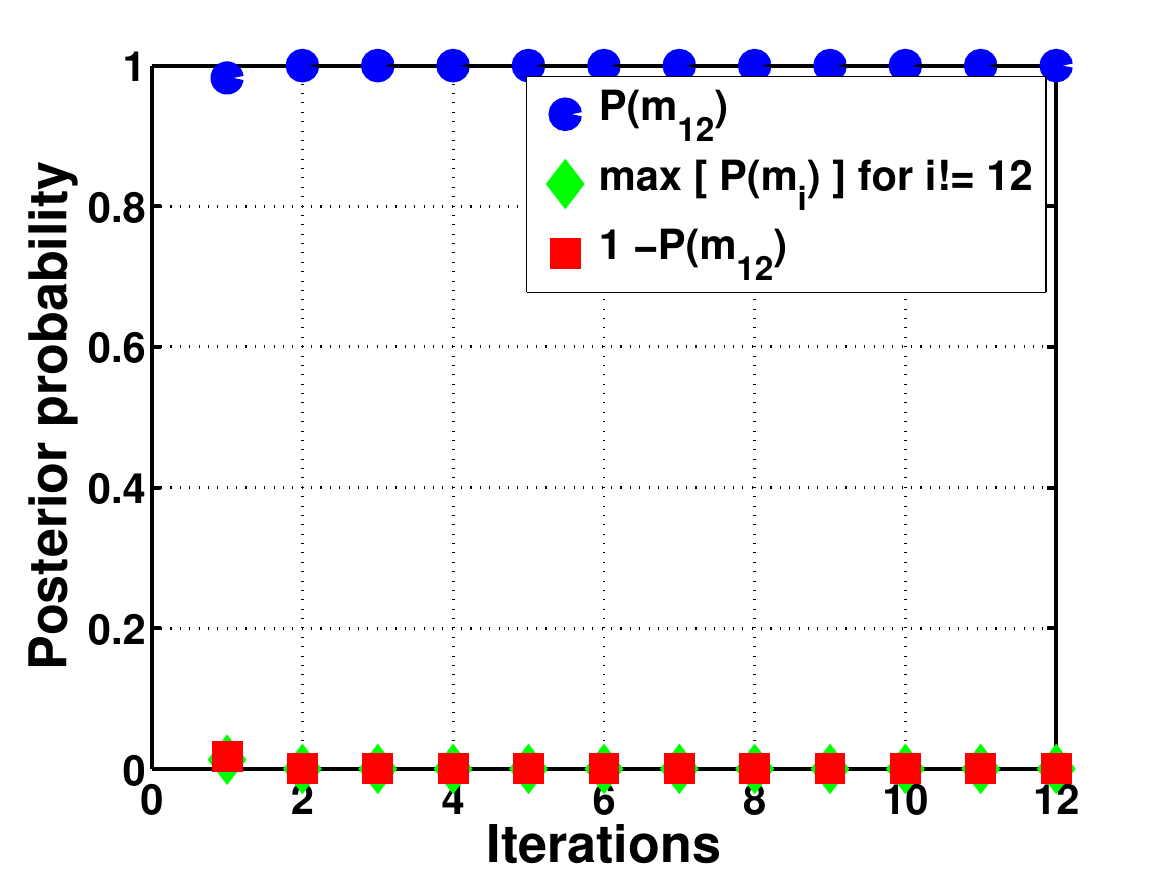}
  \caption{Multi-modal iterative estimation of material 12.}
  \label{figure:RBE2}
  \end{center}
\end{figure}

Figure~\ref{figure:RBE2} shows the evolution of the posterior
probabilities for the recognition of material 12 for the same test
sequence shown in Figure~\ref{figure:RBE}. As it can be seen
(cf. Figure~\ref{figure:RBE}) including thermal information makes
the identification correct from the first iteration on, reducing in
this case the identification time from $1$ second to $0.25$ seconds.

\section{Conclusions and Future Work}
\label{sect:conclusion}

This paper presents a multi-modal tactile based continuous material
identification approach. While state of the art approaches to material
identification mostly rely on vibration information, we show that
including thermal features reduces the material classification
errors. Moreover, standard tactile identification techniques typically
require a sequence of at least one second to classify materials. Using
recursive Bayesian estimation a robot endowed with tactile sensors can
identify materials in an average of $0.28$ seconds with a very small
deviation from that time lapse. This detection speed is again due to
the use of multi-modal information. Hence thermal sensing not only
reduces the number of errors but also enables a faster identification
than a vibration only approach.  We also eliminated the need for one
or several long exploratory movements for surface identification found
in the related literature.

Including thermal information brings the average material
identification time very close to the used window size. Faster
identification could be achieved by reducing the size of the window,
yet the selection of $\Delta t=0.25$ was empirically found to provide
an excellent time-recognition trade-off. Therefore, shorter time will
possibly imply worse accuracy. Our future work will include
other sensing modalities, specifically vision, to enhance material
recognition, for instance, generating visual texture based
priors. Tactile information could also generate priors for the visual
recognition of objects, and, in general, these two sensing modalities
could be used together in a continuous object identification system
with joint visual-tactile object models.

\bibliographystyle{IEEEtran}
\bibliography{gomez16continuous}

\begin{thebibliography}{10}
\providecommand{\url}[1]{#1}
\csname url@rmstyle\endcsname
\providecommand{\newblock}{\relax}
\providecommand{\bibinfo}[2]{#2}
\providecommand\BIBentrySTDinterwordspacing{\spaceskip=0pt\relax}
\providecommand\BIBentryALTinterwordstretchfactor{4}
\providecommand\BIBentryALTinterwordspacing{\spaceskip=\fontdimen2\font plus
\BIBentryALTinterwordstretchfactor\fontdimen3\font minus
  \fontdimen4\font\relax}
\providecommand\BIBforeignlanguage[2]{{%
\expandafter\ifx\csname l@#1\endcsname\relax
\typeout{** WARNING: IEEEtran.bst: No hyphenation pattern has been}%
\typeout{** loaded for the language `#1'. Using the pattern for}%
\typeout{** the default language instead.}%
\else
\language=\csname l@#1\endcsname
\fi
#2}}

\bibitem{Gorges.2010}
N.~Gorges, S.~Navarro, D.~G{\"o}ger, and H.W{\"o}rn, ``Haptic object
  recognition using passive joints and haptic key features,'' in \emph{Proc. of
  IEEE Intl. Conf. on Robotics and Automation}, 2010, pp. 2349--2355.

\bibitem{Scheneider.2009}
A.~Scheneider, J.~Sturm, C.~Stachniss, M.~Reisert, H.Burkhardt, and
  W.~Burkhardt, ``Object identification with tactile sensors using
  bag-of-features,'' in \emph{IEEE/RSJ Intl. Conf. on Intel. Robots and
  Systems}, 2009, pp. 243--248.

\bibitem{Chitta.2011}
S.~Chitta, J.~Sturm, M.~Piccoli, and W.~Burgard, ``Tactile sensing for mobile
  manipulation,'' \emph{IEEE Trans. on Robotics}, vol.~3, no.~27, pp. 558--568,
  2011.

\bibitem{Romano.2011}
J.~Romano, K.~Hsiao, G.~Niemeyer, S.~Chitta, and K.~Kuchenbecker,
  ``Human-inspired robotic grasp control with tactile sensing,'' \emph{IEEE
  Trans. on Robotics}, vol.~6, no.~27, pp. 1067--1079, 2011.

\bibitem{hang14hierarchical}
K.~Hang, M.~Li, J.~Stork, Y.~Bekiroglu, A.~Billard, and D.~Kragic,
  ``Hierarchical fingertip space for synthesizing adaptable fingertip grasps,''
  in \emph{ICRA 2014 Workshop: Autonomous Grasping and Manipulation: An Open
  Challenge}, 2014.

\bibitem{Edwards.2008}
J.~Edwards, J.~Lawry, and C.~Melhuish, ``Extracting textural features form
  tactile sensors,'' \emph{Bioinspiration \& Biomimetics}, vol.~3, no.~3, p.
  035002, 2008.

\bibitem{Sinapov.2011}
J.~Sinapov, V.~Sukhoy, R.~Sahai, and A.~Stoytchev, ``Vibrotactile recognition
  and categorization of surface textures by a humanoid robot,'' \emph{IEEE
  Trans. on Robotics}, vol.~3, no.~27, pp. 488--497, 2011.

\bibitem{Jamali.2011}
N.~Jamali and C.~Sammut, ``Majority voting: material classification by tactile
  sensing using surface texture,'' \emph{IEEE Trans. on Robotics}, vol.~3,
  no.~27, pp. 508--521, 2011.

\bibitem{Decherchi.2011}
S.~Decherchi, P.~Gastaldo, R.~Dahiya, M.~Valle, and R.~Zunino, ``Tactile-data
  classification of contact materials using computational intelligence,''
  \emph{IEEE Trans. on Robotics}, vol.~3, no.~27, pp. 635--639, 2011.

\bibitem{Oddo.2011}
C.~Oddo, M.~Controzzi, L.~Beccai, C.~Cipriani, and M.~Carrozza, ``Roughness
  encoding for discrimiation of surfaces in artificial active-touch,''
  \emph{IEEE Trans. on Robotics}, vol.~3, no.~27, pp. 522--533, 2011.

\bibitem{Dallaire.2013}
P.~Dallaire, P.~Gigu{\`e}re, D.~{\'E}mond, and B.~Chaib-draa, ``Autonomous
  tactile perception: A combined improved sensing and bayesian nonparametric
  approach,'' \emph{Robotics and autonomous systems}, no.~62, pp. 422--435,
  2014.

\bibitem{Chathuranga.2013}
D.~Chathuranga, V.~Ho, and S.~Hirai, ``Investigation of a biomimetic
  fingertip's ability to discriminate fabrics based on surface textures,'' in
  \emph{2013 IEEE/ASME Intl. Conf. on Advanced Intelligent Mechatronics:
  Mechatronics for Human Wellbeing}, 2013, pp. 1667--1674.

\bibitem{Chathuranga.2015}
D.~Chathuranga, Z.~Wang, Y.~Noh, T.~Nanayakkara, and S.~Hirai, ``Robust real
  time material classification algorithm using soft three axis tactile sensor:
  Evaluation of the algorithm,'' in \emph{2015 IEEE/RSJ Intl. Conf. on Intel.
  Robots and Systems}, 2015.

\bibitem{Fishel.2012}
J.~Fishel and G.~Loeb, ``Bayesian exploration for intelligent identification of
  textures,'' \emph{Frontiers in Neurorobotics}, no.~6, pp. 1--20, 2012.

\bibitem{Xu.2013}
D.~Xu, G.~Loeb, and J.~Fishel, ``Tactile identification of objects using
  bayesian exploration,'' in \emph{2013 IEEE Intl. Conf. on Robotics and
  Automation}, 2013, pp. 3056--3061.

\bibitem{Su.2012}
Z.~Su, J.~Fishel, T.~Yamamoto, and G.~Loeb, ``Use of tactile feedback to
  control exploratory movements to characterize object compliance,''
  \emph{Frontiers in Neurorobotics}, no.~6, pp. 1--9, 2012.

\bibitem{Kerr.2014}
E.~Kerr, T.~Mcginnity, and S.~Coleman, ``Material classification based on
  thermal and surface texture properties evaluated against human performance,''
  in \emph{13th Intl. Conf. on control automation robotics \& vision}, 2014,
  pp. 10--12.

\bibitem{Bhattacharjee.2015}
T.~Bhattacharjee, J.~Wade, and C.~Kemp, ``Material recognition from heat
  transfer given varying initial conditions and short-duration contact,'' in
  \emph{Proc. of Robotics: Science and Systems}, 2015.

\bibitem{BiotacManual}
J.~Fishel, G.~Lin, and G.~Loeb, ``Syntouch {LLC} biotac product manual, v.
  16,'' Tech. Rep., 2013.

\end{thebibliography}

\end{document}